%% file: main.tex
\documentclass[conference]{IEEEtran}
\IEEEoverridecommandlockouts
\usepackage{cite}
\usepackage{amsmath,amssymb,amsfonts}
\usepackage{algorithmic}
\usepackage{graphicx}
\usepackage{textcomp}
\usepackage[table,xcdraw]{xcolor}
\def\BibTeX{{\rm B\kern-.05em{\sc i\kern-.025em b}\kern-.08em
    T\kern-.1667em\lower.7ex\hbox{E}\kern-.125emX}}

\usepackage{multirow}
\usepackage{rotating}
\usepackage{enumitem} 
\usepackage{float} 
\usepackage{comment} 

\makeatletter
\def\ps@IEEEtitlepagestyle{%
\def\@oddfoot{\mycopyrightnotice}%
\def\@evenfoot{}%
}
\def\mycopyrightnotice{%
\gdef\mycopyrightnotice{}
}

\begin{document}

\title{A Parameterizable Convolution Accelerator\\
for Embedded Deep Learning Applications

}

\author{\IEEEauthorblockN{Panagiotis Mousouliotis and Georgios Keramidas}
\IEEEauthorblockA{\textit{School of Informatics}\\
\textit{Aristotle University of Thessaloniki}\\
Thessaloniki, Greece\\
\{pmousou,gkeramidas\}@csd.auth.gr}
}

\maketitle

\begin{abstract}
Convolutional neural network (CNN) accelerators implemented on Field-Programmable Gate Arrays (FPGAs) are typically designed with a primary focus on maximizing performance, often measured in giga-operations per second (GOPS). However, real-life embedded deep learning (DL) applications impose multiple constraints related to latency, power consumption, area, and cost. This work presents a hardware-software (HW/SW) co-design methodology in which a CNN accelerator is described using high-level synthesis (HLS) tools that ease the parameterization of the design, facilitating more effective optimizations across multiple design constraints. Our experimental results demonstrate that the proposed design methodology is able to outperform non-parameterized design approaches, and it can be easily extended to other types of DL applications.
\end{abstract}


\input{sections/1_introduction}
\input{sections/2_related_work}
\input{sections/3_accelerator_template}
\input{sections/4_design_parameters}
\input{sections/5_app_level_design}
\input{sections/6_experiments}
\input{sections/7_results_discussion}
\input{sections/8_conclusions_future_work}
\input{sections/9_acknowledgment}

\bibliographystyle{IEEEtran}
\bibliography{main}

\end{document}

%% file: sections/1_introduction.tex
\section{Introduction}

Embedded DL applications based on CNNs have become ubiquitous in everyday life in applications such as robotics, smartphone apps, security cameras, cars, and drones. These CNN applications introduce constraints related to execution latency, power consumption, and cost. A CNN accelerator should be designed to meet these constraints and should also be workload-aware by adjusting its design parameters either for the efficient execution of a specific CNN or a group of CNNs. Furthermore, in many DL applications, where CNNs are used as backbones for feature extraction, intermediate CNN results (called feature maps) must be readily accessible.

The most popular solution for implementing accelerated deep learning applications involves the use of graphic processing units (GPUs) and the accompanying SW frameworks. GPUs in general require hundreds of Watts for their operation and introduce complex SW requirements, which often cannot be met by resource-constrained DL applications. Embedded GPUs offer an alternative by requiring tens of Watts for their operation, but the SW requirements remain. Although GPUs present an attractive solution for use in the implementation of DL applications, their HW architecture is fixed and cannot be adapted to better support a specific CNN model or a specific CNN model compression technique.

Embedded system-on-chip (SoC) FPGAs offer all the flexibility missing from a GPU-based solution, leading to more power-efficient designs (under 10 Watts). Unfortunately, this flexibility comes at the cost of an increased design effort. Embedded SoC FPGAs consist of a host central processing unit (CPU) which is able to communicate with the FPGA fabric through standard data buses. The host CPU can be used for the development of a SW application that controls the accelerator design on the FPGA fabric. The FPGA fabric is reconfigurable; this property makes it possible to reprogram the FPGA and modify the accelerator design while the SW application executes on the host CPU. FPGA reconfiguration introduces some latency that might be acceptable depending on the application requirements.

FPGA design requires the description of the accelerator using hardware description languages (HDLs). HDLs can be used to describe an accelerator at the register-transfer level. Although HDLs offer the greatest degree of flexibility for the design of an accelerator, they increase the required design effort. To reduce design effort, FPGA vendors and academia are working on the development of tools that can be used to generate HDL code from HLS language descriptions. HLS languages \cite{cong2022fpga}, although not yet standardized, can be used to describe an FPGA design at a higher level of abstraction compared to HDLs. This allows the designer to work at the algorithmic level of abstraction, and, as with HDLs, it is possible to introduce design parameters leading to parameterized accelerator templates. These parameters can be set to build a specific accelerator that will better match the workload to be executed, leading to latency, power, and cost gains.  

This work presents the design and implementation of an HLS parameterizable convolution FPGA accelerator template that can be used in DL applications based on a CNN workload. The solution presented is based on an SoC FPGA using a HW/SW co-design methodology; the SW application is developed to run on the host CPU and invoke the FPGA accelerator. The SW application loads and preprocesses the input data and begins the execution of the CNN layers; the convolution layers are executed by the FPGA accelerator, and the rest of the layers are executed by the host CPU. 

The contributions of this work are: \textbf{i)} the presentation of the design of the convolution accelerator template, \textbf{ii)}  the description of the HLS design optimizations used, \textbf{iii)}  the presentation of system-level HW/SW co-design optimizations, \textbf{iv)}  the evaluation in terms of latency, resource usage, model accuracy, and power of the accelerator template using as workload lightweight and large CNNs \cite{iandola2016squeezenet, gschwend2016zynqnet, wang2018pelee, simonyan2014very}, and \textbf{v)}  the presentation of comparisons with related works.

The rest of this paper is organized as follows. Section \ref{sec:related_work} presents the related work in the field of FPGA accelerators for CNNs and their design methodologies. Section \ref{sec:accelerator_template} describes the architecture of the accelerator template and its parameterized design. Section \ref{sec:design_params} presents the parameters of the accelerator design. Section \ref{sec:app-level} describes application-level design characteristics and Section \ref{sec:experiments} provides information on the experimental setup and the required measurements. Section \ref{sec:results_discussion} presents the results, and, finally, Section \ref{sec:conclusions} concludes the paper.

%% file: sections/2_related_work.tex
\section{Related Work}
\label{sec:related_work}

Depending on their architecture, CNN FPGA accelerators can be classified into two groups; single computation engines \cite{guo2018angel} and streaming architectures \cite{venieris2018fpgaconvnet}.

\textit{Single Computation Engines:} A single computation engine executes each accelerated CNN layer one after another by loading the CNN layer input and the parameters each time. It is based on a design template that provides the number of processing elements (PEs) and/or the convolution loop tiling/unrolling factors as design parameters. Single computation engines can be used to accelerate many CNNs without requiring FPGA reconfiguration. In Angel-Eye \cite{guo2018angel}, \textit{Kuo et al.} present a design flow to map CNNs on embedded FPGA devices. This design flow includes a dynamic fixed-point quantization strategy, a SW-controlled HW architecture, and a run-time workflow that allows multiple CNNs to process a single frame. Since real-time processing is of main concern, Angel-Eye is designed to use a batch size of one in order to minimize latency.

\textit{Streaming Architectures:} Streaming accelerator designs provide parameterized layer kernel design templates that are connected in a dataflow fashion. Streaming accelerator architectures are tailored to a specific CNN workload and require layer execution kernels to be repeated multiple times using design parameters to match the characteristics of each layer. This leads to the usage of a large amount of FPGA resources, which results in increased power consumption and often requires FPGA reconfiguration (when the CNN graph does not fit in the FPGA), which increases the execution latency.
To overcome the limitation of FPGA reconfiguration, an alternative approach implements a streaming architecture for a CNN subgraph that is capable of running all CNN layers using a layer parameter reload mechanism \cite{venieris2018fpgaconvnet}. Streaming architectures require FPGA reconfiguration when the target CNN changes.

\textit{Algorithm/HW Co-design:} Algorithm/HW co-design architectures can be considered as a special case of the single computation engine architectures where the target CNN architecture is co-developed with the accelerator architecture. \textit{Gschwend} \cite{gschwend2016zynqnet} converted all layers, except the last global pool layer, of the SqueezeNet v1.0 CNN architecture \cite{iandola2016squeezenet} to convolutional layers. The resulting CNN, called ZynqNet, was accelerated, using an HLS floating-point design; ZynqNet describes both a CNN architecture and an HLS accelerator. 

\textit{Parameterized Single Computation Engine Architectures:} In order to improve the execution performance of a specific CNN model on a target FPGA device, single computation engine architectures are described as parameterized designs in which the number of parallel operations and the amount of on-chip (OCM) memory used are tuned to match a specific CNN architecture. Targeting a different CNN requires FPGA reconfiguration. \textit{Ma et al.} \cite{ma2018optimizing} designed an accelerator that supports convolution, fully connected, and pooling layers following a strategy that minimizes computing latency, partial sum storage, access to the on-chip buffer, and access to external memory.

This work uses HLS tools to describe a parameterized single computation engine accelerator. It is possible to set the design parameters to map CNNs to any FPGA size depending on the application requirements. This characteristic is extremely important in low-power embedded applications where power consumption is as important as execution time. It is also possible to set the design parameters to support a specific CNN or a group of CNNs (avoiding FPGA reconfiguration). Section \ref{sec:accelerator_template} describes the architecture of the parameterized accelerator template and introduces the design parameters.

%% file: sections/3_accelerator_template.tex
\section{Parameterized Accelerator Template}
\label{sec:accelerator_template}

\subsection{Supported Operations}

Considering embedded applications, the operations supported by the accelerator are the convolution, the non-linear activation function, and the max-pool. Since lightweight CNNs do not make heavy use of the fully-connected layer, it is not considered for acceleration.

\subsubsection{Convolution}

\begin{figure}
\centering
\includegraphics[scale=0.3]{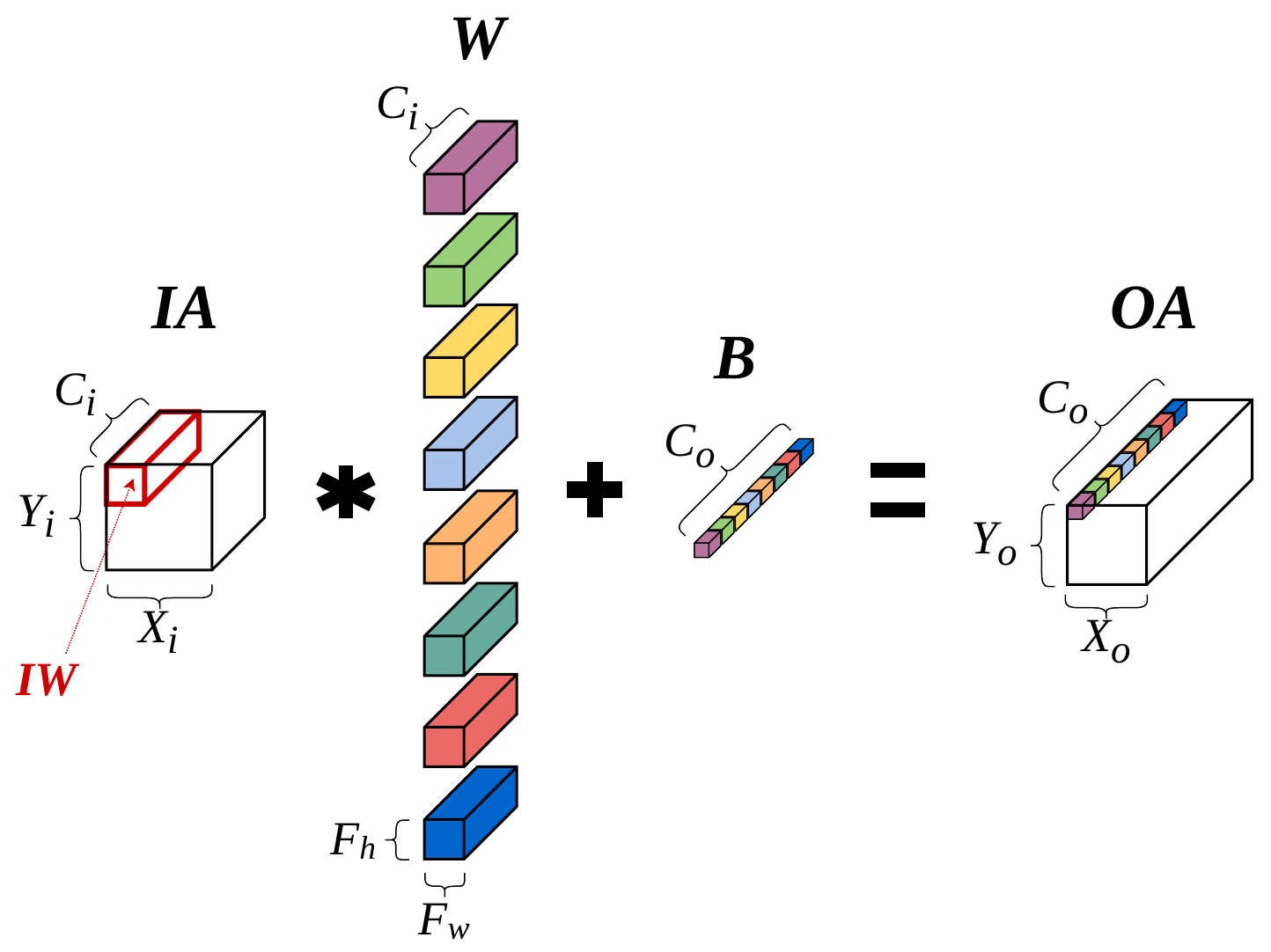}
\caption{Operation of the convolution layer. The number of 3D filters is equal to ${C_o}$, the number of output channels.}
\label{fig:conv-layer}
\vspace{-1.5em}
\end{figure}

The convolution (CONV) layer can be considered as a generalization of the 2D convolution\footnote{In the CONV layer, the convolution refers to the operation of correlation.} which is found in the classical field of image processing. \textbf{Fig.} \ref{fig:conv-layer} depicts the operation of the CONV layer in a form of a symbolic equation. On the left side of the symbolic equation, the input and parameter (weight and bias) data are shown, which are used to compute the pixels on the right side of the equation. For example, to calculate the purple channel of the 3D pixel\footnote{We define a 3D pixel as all the channels at a specific spatial location $(x,\,y)$ of the activation 3D volume.} shown on the right side of the symbolic equation, the Input Window ${IW}$ is convolved with the purple 3D filter, and the purple bias value is added to the result of the convolution. \textbf{Equation} (\ref{eq:convol}) provides the mathematical description of the operation of the CONV layer.
\begin{equation}
\label{eq:convol}
\begin{gathered}
{OA}(y_o ,\, x_o ,\, c_o)=\\
\sum_{f_h=0}^{{F_h}-1}\sum_{f_w=0}^{{F_w}-1}\sum_{c_i=0}^{{C_i}-1}\{
{IW} \cdot {W}(c_o ,\ f_h ,\ f_w ,\ c_i)\}
+ {B}(c_o)
\end{gathered}
\end{equation}
where ${IW} = {IA}(y_o \cdot {S}+f_h,\ x_o \cdot {S}+f_w,\ c_i)$, ${OA}$ and ${IA}$ are the output and the input activations respectively, and ${W}$, ${B}$ are the weight and bias parameters respectively. The $y$, $x$, and $c$, represent the vertical, horizontal, and channel dimensions of the activations, ${S}$ is the stride, and $f_h$, $f_w$ are the vertical and horizontal dimensions of the filters. The sum in (\ref{eq:convol}) represents the convolution between ${IW}$ and the ${c_o}$ 3D filter. This value is then added to the respective bias term ${B}(c_o)$ to produce the output channel value of the output pixel of ${OA}$. To calculate all the pixels of ${OA}$, ${IW}$ is shifted horizontally and vertically by $x_o \cdot {S}$ and $y_o \cdot {S}$ respectively, and the aforementioned procedure is repeated.

\subsubsection{Optional Activation Layer and Max-Pool}

The rectified linear unit (ReLU) layer \cite{krizhevsky2012imagenet} is embedded in the convolution operation due to its convenient element-wise operation; it can be optionally deactivated to support convolution layers not followed by activation layers. The HLS accelerator description allows ReLU to be easily replaced by any activation function. The ReLU activation function is described by $f(x)=max(0, \, x)$. The CONV operation is followed by an optional execution of the max-pool (MPOOL) operation; MPOOL calculates the maximum value on a sliding window on each activation channel.

\subsubsection{Summary}

\textbf{Equation} (\ref{eq:accel_op}) summarizes the accelerator operation and the final output (the \textcolor{blue}{blue} color marks the optional usage of the ReLU and MPOOL operations). The accelerator supports the following layer types: convolutions with filter size $1 \times 1$ (padding 0), filter size $3 \times 3$ (padding 1 or 0, stride 1 or 2), optional ReLU, optional $3 \times 3$ max-pooling with stride 2 and $2 \times 2$ max-pooling with stride 2. The accelerator's operation can be easily adapted to support additional layers and filter sizes since it is described using HLS tools.
\begin{equation}
\label{eq:accel_op}
\textsf{OACT} = \textcolor{blue}{\textsf{MPOOL}(\ \textsf{ReLU}(}\ \textsf{CONV}(IA,\,W,\,B)\ \textcolor{blue}{)\ )}
\end{equation}

\subsection{Architecture}

\begin{figure}
\centering
\includegraphics[scale=0.33]{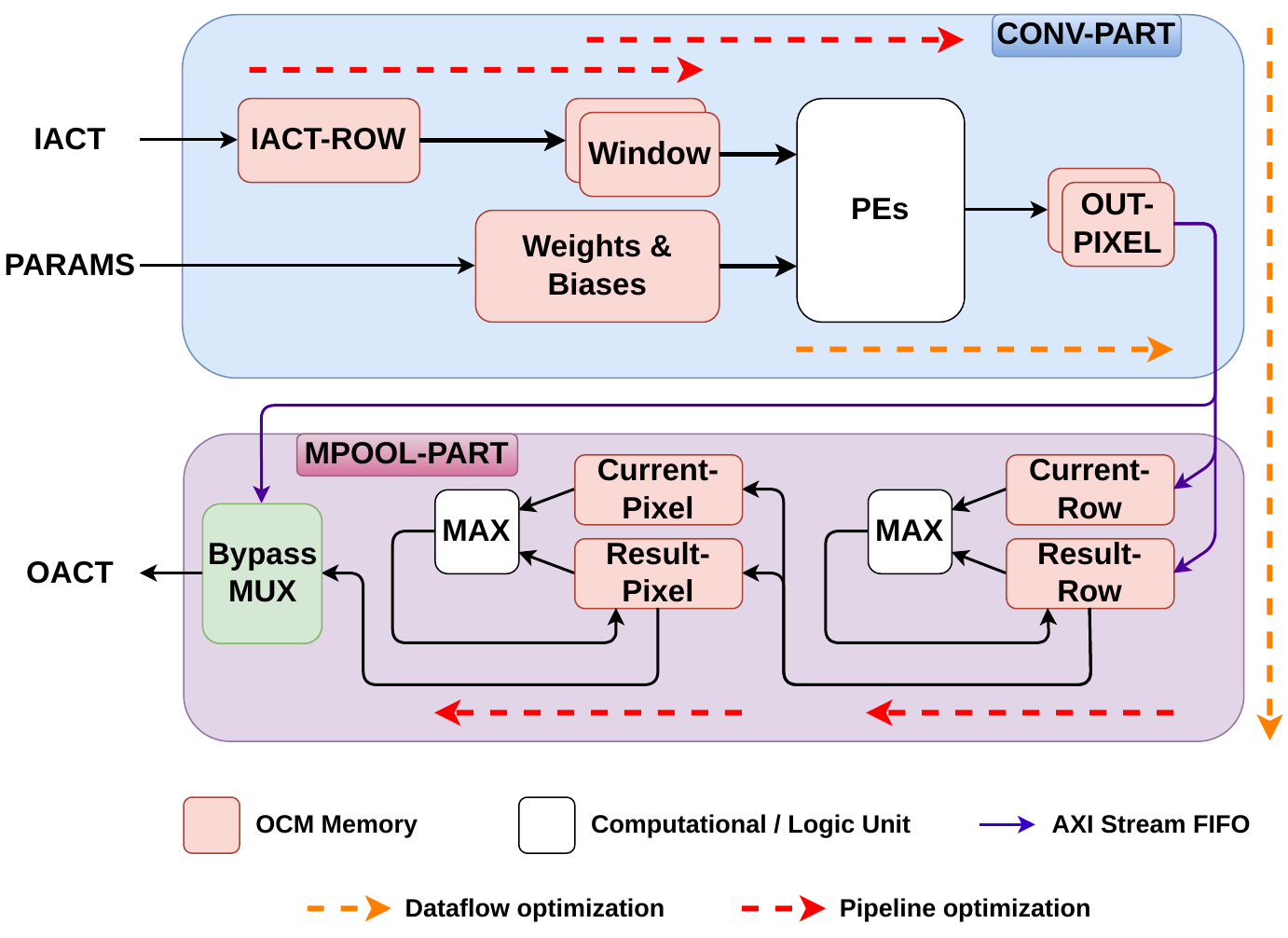}
\caption{Accelerator block diagram augmented with HLS optimizations.}
\label{fig:accel}
\vspace{-1em}
\end{figure}

\textbf{Fig.} \ref{fig:accel} shows the accelerator block diagram. Inputs to the accelerator are the input activation values (IACT), the CNN weights and biases (PARAMS), and the scalar inputs, such as the filter size, the stride, the padding, etc., which are omitted for simplicity. The accelerator output consists of the output activation values (OACT) calculated by (\ref{eq:accel_op}). As depicted in \textbf{Fig.} \ref{fig:accel}, the accelerator consists of two parts; the CONV-PART and the MPOOL-PART. The CONV-PART implements the CONV and ReLU operations of (\ref{eq:accel_op}). The output of the CONV-PART is the input of the MPOOL-PART; the MPOOL-PART will either perform the MPOOL operation and pass the result to the accelerator output, or it will bypass the MPOOL operation and pass to the accelerator output the CONV-PART result.

\subsubsection{Convolution Part}

The IACT data entering the accelerator block is organized as $[H_i,\, X_i,\, C_i]$ with $C_i$ being the fastest-changing dimension. The reasons for this selection are: i) during the CONV operation, the calculation of one output channel requires all input channels to be present (see \textbf{Fig.} \ref{fig:conv-layer}), and ii) if the PARAMS values fit in the OCM memory, the IACT is required to be streamed only once to the accelerator, achieving optimal data transfer. Unfortunately, low-end SoC FPGAs cannot always hold all the CONV layer parameters on-chip, even when data quantization takes place. In this case, the CONV parameters, whose data are organized as $[C_o,\, F_h,\, F_w,\, C_i]$, are split in the $C_o$ dimension and the original convolution is split into secondary convolutions whose results are finally merged to acquire the result of the original convolution. This split-merge operation requires the IACT to be re-streamed in the accelerator as many times as the number of the secondary convolutions formed. The structure of the accelerator CONV-PART follows the structure of the symbolic equation in \textbf{Fig.} \ref{fig:conv-layer}. The IACT-ROW is filled with 3D IACT rows, the Window OCM is the IW of the figure, the Weights \& Biases OCMs hold the CONV parameters, the PEs perform the calculations, and the results are written in the OUT-PIXEL OCMs.

\subsubsection{Max-pool Part}

\begin{figure}
\centering
\includegraphics[scale=0.49]{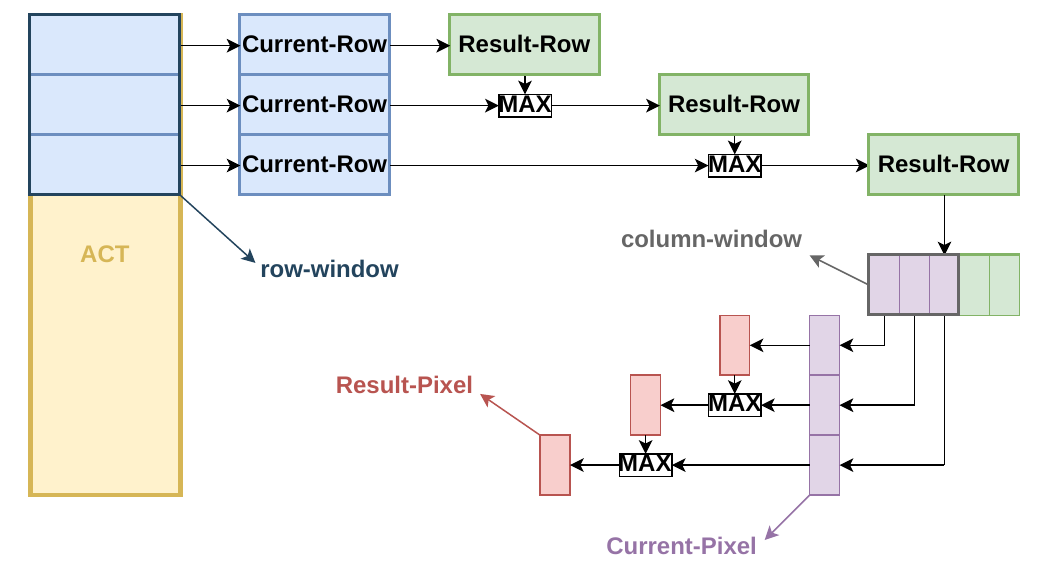}
\caption{The accelerator implementation of the MPOOL operation for the case of a $3 \times 3$ filter. The channel dimension is omitted for simplicity.}
\label{fig:accel-mpool-op}
\vspace{-1.5em}
\end{figure}

\textbf{Fig.} \ref{fig:accel-mpool-op} shows the max-pool operation of the accelerator. Although the channel dimension is omitted for clarity, it must be stressed that it is the parallelism in the channel dimension that is exploited to accelerate this operation. Specifically, each MAX calculation in the figure performs spatial MAX operations for multiple channels in parallel. These MAX operations are performed between 3D rows in the height dimension, using Current-Row and Result-Row, and, subsequently, between 3D pixels in the width dimension, using Current-Pixel and Result-Pixel. 

\subsection{Quantization Strategy}

The use of floating-point values for CNN layer execution requires FPGA OCM sizes that cannot be found in low-end SoC FPGAs suitable for low-power embedded applications. Furthermore, floating-point operations require a considerable amount of FPGA resources; this decreases the opportunities for parallelism and increases the power consumption (when floating-point hard cores are not present). For the aforementioned reasons, the accelerator is configured to use 8-bit operands for CNN activation and parameter data. These operands follow the dynamic fixed-point (DFP) quantization format \cite{gysel2018ristretto} that allows the use of 8-bit data with negligible loss of CNN accuracy.

\subsection{Parallelism Exploitation using HLS}  

\begin{figure}
\centering
\includegraphics[scale=0.5]{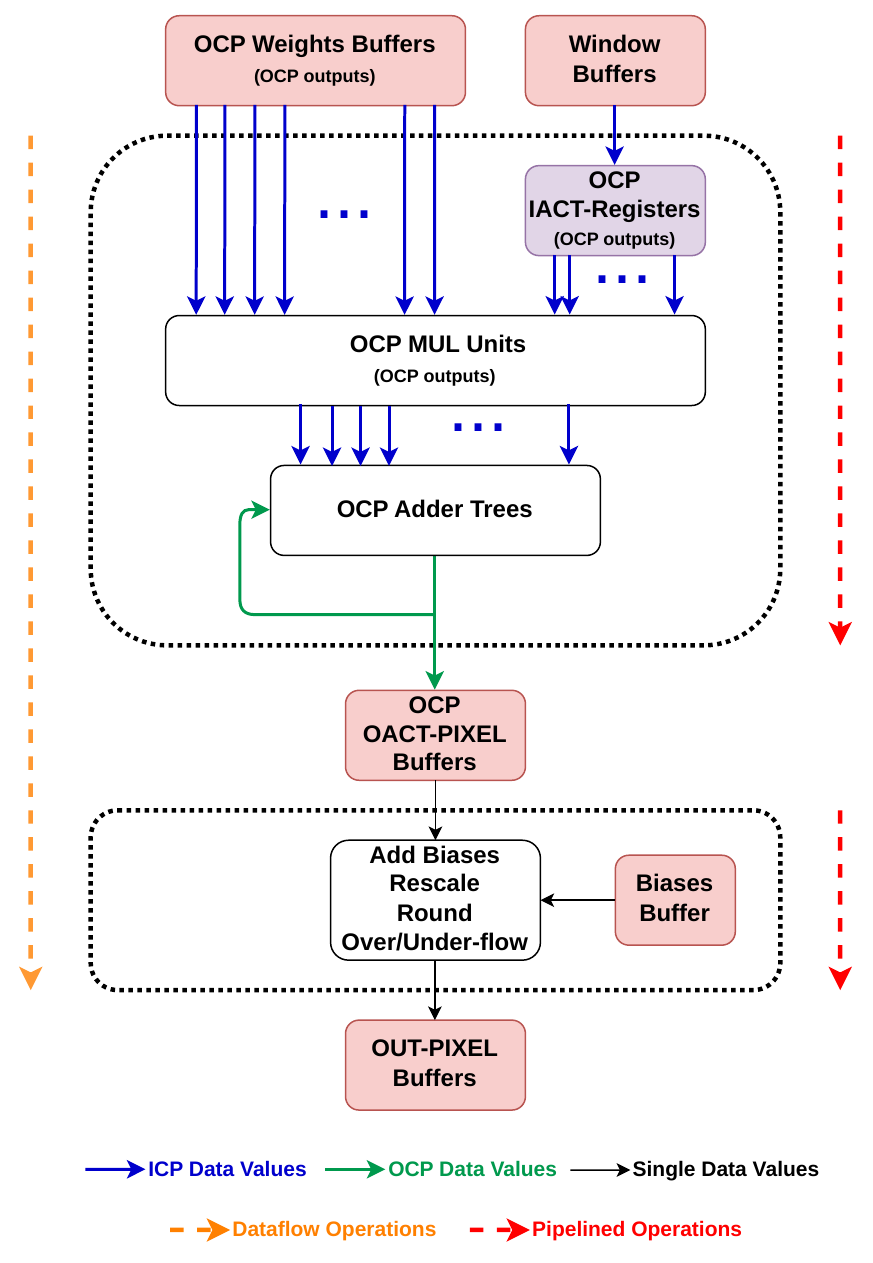}
\caption{The implementation of the CONV operation.}
\label{fig:accel-conv}
\vspace{-1.5em}
\end{figure}

The accelerator is described in C code and the HLS optimizations are introduced as pragmas for use by the HLS synthesis tool. At the top design level, data-transfer parallelism is implemented using the \texttt{data\_pack} optimization. The data is packed at the host processor of the SoC FPGA in order to take advantage of the width of the high-performance ports used to transfer the data to the FPGA accelerator and vice versa. As it is shown in \textbf{Fig.} \ref{fig:accel}, a \texttt{dataflow} optimization is used for pipelining the operation of the CONV-PART and the MPOOL-PART; the dataflow HLS optimization is used for task-level temporal parallelism. The CONV-PART calculates $C_o$ number of values which are consumed by the MPOOL-PART as they become available. The \texttt{stream} HLS optimization is used to signify that the data transfer between these two modules is done in a way that allows them to operate in a pipelined fashion. A similar optimization used for temporal parallelism is the \texttt{pipeline} HLS optimization which is used for pipelining simple operations, the data transfers between the host CPU and the accelerator, and the data transfers between the accelerator OCMs.

The \texttt{array\_partition} HLS optimization is heavily used in both the CONV-PART and the MPOOL-PART of the accelerator. This optimization organizes the data stored in the FPGA RAM resources in order to provide multiple ports for reading and writing data; this is a form of spatial parallelism optimization. For example, the IACT-ROW, Window, and Weights OCMs of the CONV-PART are partitioned with an input channel parallelism (ICP) factor providing ICP number of parallel ports for reading and writing data. Additionally, the Weights \& Biases OCMs are partitioned into output channel parallelism (OCP) number of FPGA RAM resources. The reason is to provide in parallel values to the OCP parallel operating PEs that implement the MAC operation. In this way, OCP number of PEs perform ICP number of parallel multiplications each. Each PE also has an adder tree structure for adding the intermediate results for the implementation of the accumulation operation. Finally, $C_o$ CONV results are written in the OUT-PIXEL OCM and are forwarded to the MPOOL-PART. \textbf{Fig.} \ref{fig:accel-conv} shows the implementation of the CONV operation augmented with HLS optimizations.

In order to overlap data communication and computation, the accelerator CONV-PART incorporates a double buffering technique. Specifically, it includes two Window and two OUT-PIXEL OCMs  for this purpose (see \textbf{Fig.} \ref{fig:accel}).

Two additional optimizations used in the CONV calculation are the squeeze of two multiplication operations into one DSP FPGA block and the use of the \texttt{resource} HLS pragma for selecting a number of the OCP PEs to be implemented using LUT resources instead of DSP resources. In this way DSP resources can be saved or even do not be used at all depending on the FPGA resource configuration.

The accelerator CONV-PART takes advantage of the $C_i$ and $C_o$ channel dimensions to introduce parallelism. Similarly, the MPOOL-PART uses parallelism in the channel dimension. The MAX operation is performed in multiple pixel channels in parallel using row and column 3D windows defined by the MPOOL filter size (see \textbf{Fig.} \ref{fig:accel-mpool-op}).

%% file: sections/4_design_parameters.tex
\section{Design Parameters}
\label{sec:design_params}

\begin{table}
\centering
\caption{Accelerator Design Parameters}
\label{tab:accel-design-params}
\scalebox{0.85}{
\begin{tabular}{c|ll}
\hline
\textbf{TOP} & \multicolumn{2}{l}{\cellcolor[HTML]{EFEFEF}\begin{tabular}[c]{@{}l@{}}\textbf{FREQ}\\ Accelerator operating frequency.\end{tabular}} \\
 & \multicolumn{2}{l}{\begin{tabular}[c]{@{}l@{}}\textbf{APACK}\\ Number of activation values packed in one data transfer.\\ Channel parallelism factor for the MPOOL-PART.\end{tabular}} \\
 & \multicolumn{2}{l}{\cellcolor[HTML]{EFEFEF}\begin{tabular}[c]{@{}l@{}}\textbf{PPACK}\\ Number of CONV weight values packed in one data transfer.\end{tabular}} \\
 & \multicolumn{2}{l}{\begin{tabular}[c]{@{}l@{}}\textbf{ICP}\\ Input channel parallelism used in the CONV-PART.\end{tabular}} \\
 & \multicolumn{2}{l}{\cellcolor[HTML]{EFEFEF}\begin{tabular}[c]{@{}l@{}}\textbf{OCP}\\ Output channel parallelism used in the CONV-PART.\\ Equal to the number of PEs.\end{tabular}} \\
 & \multicolumn{2}{l}{\begin{tabular}[c]{@{}l@{}}\textbf{PE\_DSP}\\ Number of PEs implemented with DSP blocks.\\ The rest are implemented using LUTs.\end{tabular}} \\ \hline
\textbf{CONV} & \multicolumn{2}{l}{\cellcolor[HTML]{EFEFEF}\begin{tabular}[c]{@{}l@{}}\textbf{FILTER\_MAX}\\ Maximum size of the CONV filter.\\ Maximum number of IACT-ROW rows.\end{tabular}} \\
 & \multicolumn{2}{l}{\begin{tabular}[c]{@{}l@{}}\textbf{WINxCHIN\_PAD\_MAX}\\ Maximum size of the row of the IACT-ROW OCM.\end{tabular}} \\
 & \multicolumn{2}{l}{\cellcolor[HTML]{EFEFEF}\begin{tabular}[c]{@{}l@{}}\textbf{FILTERxFILTERxCHIN\_MAX}\\ Maximum size of the Window OCMs.\end{tabular}} \\
 & \multicolumn{2}{l}{\begin{tabular}[c]{@{}l@{}}\textbf{CHOUTxFILTERxFILTERxCHIN\_MAX}\\ Maximum size of the Weights OCMs.\end{tabular}} \\
 & \multicolumn{2}{l}{\cellcolor[HTML]{EFEFEF}\begin{tabular}[c]{@{}l@{}}\textbf{CHOUT\_MAX}\\ Maximum size of the OUT-PIXEL and\\ total size of the Biases OCMs.\end{tabular}} \\ \hline
\textbf{MPOOL} & \multicolumn{2}{l}{\begin{tabular}[c]{@{}l@{}}\textbf{PWINxPCH\_MAX}\\ Maximum size of the CURRENT/RESULT-ROW OCMs.\end{tabular}} \\
 & \multicolumn{2}{l}{\cellcolor[HTML]{EFEFEF}\begin{tabular}[c]{@{}l@{}}\textbf{PCH\_MAX}\\ Maximum size of the CURRENT/RESULT-PIXEL OCMs.\end{tabular}} \\ \hline
\end{tabular}
}
\vspace{-1.5em}
\end{table}

\textbf{Table} \ref{tab:accel-design-params} describes the accelerator design parameters. From top to bottom, the table is divided into three parameter groups. The first group is related to the top-level design, the second is related to the accelerator CONV-PART, and the third is related to the MPOOL-PART. The parameters of the top-level group deal with the accelerator system ports width (APACK, PPACK), the number of concurrent multiplications performed by one PE (ICP), the number of PEs (OCP), and the number of PEs implemented using DSP blocks (PE\_DSP). The CONV-PART and the MPOOL-PART related parameters depend on the size of the CONV and MPOOL layers that will be supported by the accelerator; these parameters are related to the OCM sizes. In order to support multiple CNNs without reconfiguring the FPGA, the CONV-PART and MPOOL-PART parameters must support the execution of the related layers of all the target CNNs. With the use of the HLS tools, modifying the above parameters is as simple as editing a C header file.

%% file: sections/5_app_level_design.tex
\section{Application-level Design}
\label{sec:app-level}

The high-performance ports used for transferring data from the host CPU to the FPGA fabric are configured to be cache-coherent. This design configuration allows the CPU to execute all CNN layers not supported by the accelerator without slowing down the application execution due to cache misses.

To overcome the reduced input channel issue of the CNN first layer \cite{gholami2018squeezenext} and to increase the computational efficiency of the accelerator template, which takes advantage of the parallelism in the input and output channel dimensions, the operation of the first CNN layer is always reshaped \cite{wu2019high}. Layer reshaping is a software solution that reorganizes the input and parameters of the first CNN layer to increase the utilization of the accelerator computation.

%% file: sections/6_experiments.tex
\section{Experiments}
\label{sec:experiments}

A C++ application is developed to run on the host CPU of the SoC FPGA and control the accelerator execution. This application is used for measuring the CNN accuracy, and the CONV and overall CNN latencies. Power consumption is estimated by the Xilinx tools in the implemented design reports. For all the experiments, the Xilinx XCZU3EG SoC FPGA is used. The accelerator is designed to operate with a batch size equal to one.

%% file: sections/7_results_discussion.tex
\section{Results and Discussion}
\label{sec:results_discussion}

\begin{table}
\centering
\caption{The characteristics of the accelerator workload}
\label{tab:cnn-workload}
\scalebox{0.9}{
\begin{tabular}{crrr}
\hline
\textbf{MODEL} &
\textbf{PARAM {[}M{]}} &
\textbf{MAC {[}M{]}} &
\textbf{ACTIV {[}M{]}}\\ \hline
SqueezeNet v1.1 & 1.2 & 352   & 7.0 \\
ZynqNet         & 2.5 & 529   & 8.6 \\
PeleeNet        & 2.8 & 514   & 12.4 \\
VGG-16          & 138.0 & 15470 & 29.0 \\ \hline
\end{tabular}
}
\vspace{-1em}
\end{table}

\begin{table}
\caption{Top-1 / top-5 accuracy of accelerator workload}
\label{tab:cnn-accuracy}
\centering
\scalebox{0.9}{
\begin{tabular}{cccc}
\hline
\textbf{CNN} & \textbf{FLP} & \textbf{DFP} & \textbf{DFP–FPGA} \\ \hline
SqueezeNet v1.1 & 58.39 / 81.00 & 57.42 / 80.22 & 57.27 / 80.14 \\
ZynqNet & 62.01 / 84.18 & 61.16 / 83.54 & 61.15 / 83.57 \\
PeleeNet & 71.20 / 90.46 & 68.11 / 88.86 & 67.46 / 88.61 \\
VGG-16 & 70.14 / 89.77 & 67.67 / 88.61 & 67.40 / 88.01 \\ \hline
\end{tabular}
}
\vspace{-1em}
\end{table}

\begin{table*}
\caption{Implementations of the Accelerator Template Configurations}
\label{tab:impl-configs}
\centering
\scalebox{0.9}{
\begin{tabular}{lrrrrrrc}
\hline
\textbf{CONF} & \multicolumn{1}{c}{\#1} & \multicolumn{1}{c}{\#2} & \multicolumn{1}{c}{\#3} & \multicolumn{1}{c}{\#4} & \multicolumn{1}{c}{\#5} & \multicolumn{1}{c}{\#6} & \cellcolor[HTML]{DAE8FC}\#6P \\ \hline
\textbf{FREQ (MHz)} & \multicolumn{1}{c}{100} & \multicolumn{1}{c}{100} & \multicolumn{1}{c}{100} & \multicolumn{1}{c}{100} & \multicolumn{1}{c}{\cellcolor[HTML]{DAE8FC}200} & \multicolumn{1}{c}{\cellcolor[HTML]{DAE8FC}300} & 300 \\
\textbf{ICP} & \multicolumn{1}{c}{16} & \multicolumn{1}{c}{16} & \multicolumn{1}{c}{16} & \multicolumn{1}{c}{\cellcolor[HTML]{DAE8FC}32} & \multicolumn{1}{c}{32} & \multicolumn{1}{c}{32} & 32 \\
\textbf{OCP} & \multicolumn{1}{c}{8} & \multicolumn{1}{c}{\cellcolor[HTML]{DAE8FC}16} & \multicolumn{1}{c}{16} & \multicolumn{1}{c}{16} & \multicolumn{1}{c}{16} & \multicolumn{1}{c}{16} & 16 \\
\textbf{PACK} & \multicolumn{1}{c}{8} & \multicolumn{1}{c}{8} & \multicolumn{1}{c}{\cellcolor[HTML]{DAE8FC}16} & \multicolumn{1}{c}{16} & \multicolumn{1}{c}{16} & \multicolumn{1}{c}{16} & 16 \\ \hline
\textbf{LUT (\%)} & 26013 (36.87) & 37300 (52.86) & 37784 (53.55) & 47440 (67.23) & 48697 (69.02) & 48434 (68.64) & \multicolumn{1}{r}{46639 (66.10)} \\
\textbf{LUTRAM (\%)} & 3288 (11.42) & 3864 (13.42) & 3864 (13.42) & 4184 (14.53) & 4485 (15.57) & 4751 (16.50) & \multicolumn{1}{r}{3203 (11.12)} \\
\textbf{FF (\%)} & 18852 (13.36) & 21058 (14.92) & 21997 (15.59) & 26528 (18.80) & 36481 (25.85) & 44926 (31.84) & \multicolumn{1}{r}{44750 (31.71)} \\
\textbf{BRAM (\%)} & 113 (52.31) & 109 (50.46) & 123 (56.94) & 187 (86.57) & 187 (86.57) & 195 (90.28) & \multicolumn{1}{r}{171 (79.17)} \\
\textbf{DSP (\%)} & 74 (20.56) & 138 (38.33) & 138 (38.33) & 266 (73.89) & 266 (73.89) & 266 (73.89) & \multicolumn{1}{r}{265 (73.61)} \\ \hline
\textbf{POWER (W)} & 2.418 & 2.463 & 2.481 & 2.710 & 3.506 & 4.259 & \multicolumn{1}{r}{3.938} \\ \hline
\textbf{SQN CONV (ms)} & 41.541 & 26.856 & 24.831 & 22.096 & 14.280 & 12.529 & - \\
\textbf{ZQN CONV (ms)} & 53.566 & 34.136 & 32.530 & 30.327 & 20.689 & 18.702 & - \\
\textbf{PLN CONV (ms)} & 62.769 & 49.768 & 47.217 & 45.725 & 30.179 & 28.071 & \multicolumn{1}{r}{27.707} \\
\textbf{VGG CONV (ms)} & 1251.248 & 651.600 & 648.315 & 347.787 & 179.896 & 126.520 & - \\ \hline
\textbf{SQN (ms)} & 47.517 & 32.840 & 30.814 & 28.119 & 20.330 & 18.557 & - \\
\textbf{ZQN (ms)} & 62.690 & 43.238 & 41.619 & 39.533 & 29.906 & 27.932 & - \\
\textbf{PLN (ms)} & 75.123 & 62.037 & 59.516 & 58.153 & 42.678 & 40.511 & \multicolumn{1}{r}{40.147} \\
\textbf{VGG (ms)} & 1421.337 & 821.675 & 818.449 & 517.957 & 349.941 & 296.807 & - \\ \hline
\end{tabular}
}
\vspace{-1em}
\end{table*}

\begin{table}
\caption{Comparison to Related Work for VGG-16 Acceleration}
\label{tab:comparison}
\centering
\scalebox{0.76}{
\begin{tabular}{lrrrr}
\hline
 & \textbf{Ma et al. \cite{ma2018optimizing}} & \textbf{Angel-Eye \cite{guo2018angel, venieris2018toolflows}} & \textbf{fpgaConvNet \cite{venieris2018toolflows}} & \textbf{This Work} \\ \hline
\textbf{FPGA} & STRATIX V & XC7Z045 & XC7Z045 & XCZU3EG \\
\textbf{FREQ (MHz)} & 150 & 150 & 125 & 300 \\
\textbf{PRECISION} & 16 bits & 16 bits & 16 bits & 8 bits \\
\textbf{CONV (ms)} & 72.50 & 163.42 & 249.50 & 126.52 \\
\textbf{LUT (\%)} & \multicolumn{1}{r}{218K ALM (93)} & \multicolumn{1}{r}{182616 (84)} & 218600 (100) & \multicolumn{1}{r}{48434 (69)} \\
\textbf{FF (\%)} & - & \multicolumn{1}{r}{127653 (29)} & - & \multicolumn{1}{r}{44926 (32)} \\
\textbf{BRAM (\%)} & \multicolumn{1}{r}{2210 M20K (86)} & \multicolumn{1}{r}{486 (89)} & 545 (100) & \multicolumn{1}{r}{195 (90)} \\
\textbf{DSP (\%)} & \multicolumn{1}{r}{256 (100)} & \multicolumn{1}{r}{780 (87)} & 900 (100) & \multicolumn{1}{r}{266 (74)} \\
\textbf{POWER (W)} & - & \multicolumn{1}{r}{9.63} & - & \multicolumn{1}{r}{4.26} \\ \hline
\end{tabular}
}
\vspace{-1em}
\end{table}

\textbf{Table} \ref{tab:cnn-workload} shows the characteristics of the CNN workload in terms of parameter count (M stands for Millions), MAC operations, and activation data count. \textbf{Table} \ref{tab:cnn-accuracy} shows the accuracy results of the CNN workload on the ImageNet 2012 validation set \cite{russakovsky2015imagenet}. FLP refers to 32-bit floating-point model parameters and activations, DFP refers to 8-bit dynamic fixed-point model parameters and activations, and DFP-FPGA refers to the 8-bit DFP results acquired when the CNN workload is executed by the FPGA accelerator on the SoC FPGA board.

\textbf{Table} \ref{tab:impl-configs} shows the implementations of various accelerator template configurations. (\%) indicates the percent resource usage for the specific FPGA device. SQN, ZQN, PLN, and VGG, are abbreviations for the CNN workload (see \textbf{Table} \ref{tab:cnn-workload}). When a CNN abbreviation is followed by CONV it refers to the CNN total CONV latency; otherwise it refers to the total end-to-end CNN latency. The shaded values indicate the tuned parameter compared to the parameter of the previous column. These results show that the implementations of the parameterized accelerator design cover a wide range of resources, power, and latency requirements. Increased parallelism (ICP, OCP) translates to resource usage increase and latency decrease. Configurations \#1 to \#4  can be used to target low-cost low-power embedded DL applications. As implementation frequency increases (CONF \#5 and \#6), power consumption increases and execution latency decreases; the resource usage slightly increases to support the additional pipeline stages required by the increased frequency. An almost 800mW of power increase per 100MHz frequency increase can be observed. For high-performance embedded DL applications, CONF \#6 can be used as it provides low CNN latencies (above 20 frames per second for all CNNs except VGG) at under 4.3Watts. Configuration \#6P is the implementation of CONF \#6 tuned for the OCM sizes of the PLN CNN. CONF \#6P presents 300mW of power improvement as well as a 11\% BRAM usage decrease.

\textbf{Table} \ref{tab:comparison} compares CONF \#6 with related works. In terms of CONV latency, CONF \#6 outperforms Angel-Eye \cite{guo2018angel, venieris2018toolflows} and fpgaConvNet \cite{venieris2018toolflows} using a lower resource count and reporting lower power consumption. \textit{Ma et al.} \cite{ma2018optimizing} outperforms CONF \#6 in terms of CONV latency, but uses a higher resource count. Quantization in lower bit widths is key to achieving reduced resource usage, reduced power consumption, and increased performance in terms of latency.

%% file: sections/8_conclusions_future_work.tex
\section{Conclusions}
\label{sec:conclusions}

This work presented the design of a parameterizable convolution FPGA accelerator architecture using HLS tools. The evaluation of the design proved the accelerator template flexibility to describe design configurations with a broad range of resources, power, and latency combinations. These design points can be used to cover low-end to more demanding embedded DL applications. Future work includes design modeling and multiobjective optimization techniques towards the automation of finding a suitable design point as required by the embedded DL application.

%% file: sections/9_acknowledgment.tex
\section*{Acknowledgment}
\label{sec:acknowledgment}
This work is part of R-PODID project, supported by the Chips Joint Undertaking and its members, including the top-up funding by National Authorities of Italy, Turkey, Portugal, The Netherlands, Czech Republic, Latvia, Greece, and Romania under grant agreement No 101112338.